\documentclass[10pt,twocolumn,letterpaper]{article}

\usepackage{cvpr}
\usepackage{times}
\usepackage{epsfig}
\usepackage{graphicx}
\usepackage{amsmath}
\usepackage{amssymb}
\usepackage{multirow}

\begin{document}

\title{Discerning the Chaos: Detecting Adversarial Perturbations while Disentangling Intentional from Unintentional Noises}

\author{Anubhooti Jain, Susim Roy, Kwanit Gupta, Mayank Vatsa, and Richa Singh\\
IIT Jodhpur\\
{\tt\small $\{$jain.44, roy.10, gupta.45, mvatsa, richa$\}$@iitj.ac.in}
}


\maketitle
\thispagestyle{empty}

\begin{abstract} 
    Deep learning models, such as those used for face recognition and attribute prediction, are susceptible to manipulations like adversarial noise and unintentional noise, including Gaussian and impulse noise. This paper introduces CIAI, a Class-Independent Adversarial Intent detection network built on a modified vision transformer with detection layers. CIAI employs a novel loss function that combines Maximum Mean Discrepancy and Center Loss to detect both intentional (adversarial attacks) and unintentional noise, regardless of the image class. It is trained in a multi-step fashion. We also introduce the aspect of intent during detection that can act as an added layer of security. We further showcase the performance of our proposed detector on CelebA, CelebA-HQ, LFW, AgeDB, and CIFAR-10 datasets. Our detector is able to detect both intentional (like FGSM, PGD, and DeepFool) and unintentional (like Gaussian and Salt \& Pepper noises) perturbations.
    
\end{abstract}

\begin{figure}[t]
\begin{center}
   \includegraphics[width=1.0\linewidth]{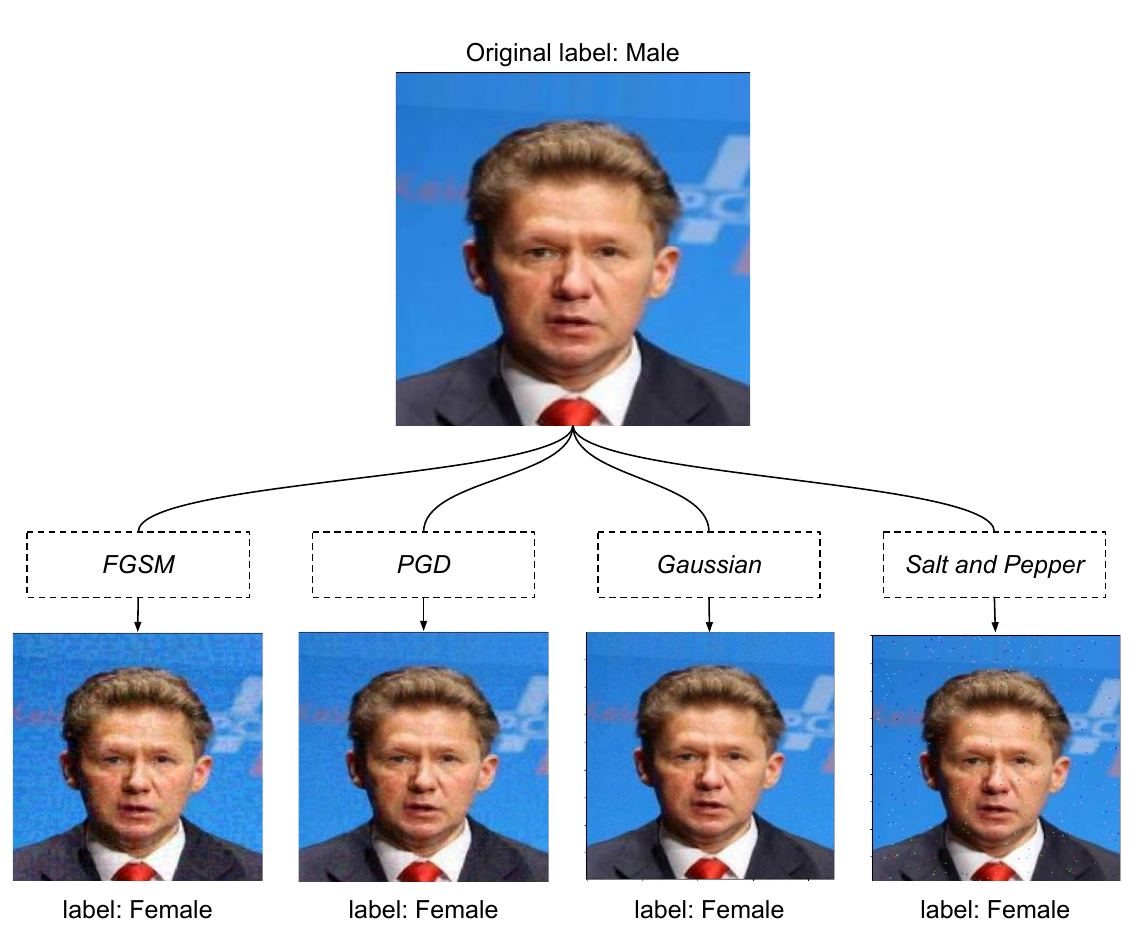}
\end{center}
   \caption{Labels affected using intentional (adversarial perturbations) as well as unintentional noises (corruptions).}
\label{fig:abs}
\end{figure}

\section{Introduction}
Adversarial Attacks \cite{DBLP:journals/caaitrit/ChakrabortyADCM21,DBLP:journals/corr/abs-2206-08304} have been a well-posed threat against deep neural networks for a long time now. For different tasks, datasets, and architectures, the attacks are a serious security issue, even when the attacked images appear normal to human eyes. 
Different attacks have been proposed over the years that can be broadly classified as white-box, gray-box, black-box, and physical adversarial attacks \cite{DBLP:conf/sp/Carlini017,DBLP:journals/corr/GoodfellowSS14,DBLP:conf/iclr/MadryMSTV18}. Several defense techniques have been proposed in order to defend against these attacks, like adversarial training, distillation, and feature squeezing \cite{DBLP:conf/cvpr/ShrivastavaPTSW17,DBLP:conf/ndss/Xu0Q18}. Some of them are computationally heavy and some are class-dependent as well. One other branch of these defensive techniques lies in detecting the attacked images so they can be caught even before being sent to the model network. While some of these methods have shown impressive accuracies, not many highlight the effect on the models' performance on unseen attacks. Also, some of these methods do not work for attacks like CW \cite{DBLP:conf/sp/Carlini017} or DeepFool \cite{DBLP:conf/cvpr/Moosavi-Dezfooli16}. 

Further, there are some noises or corruptions that can be added during image processing, such as blurring or pixelation, and are considered important to increase classifier stability \cite{DBLP:journals/corr/DodgeK17a,DBLP:journals/corr/abs-2204-13653,DBLP:journals/corr/VasiljevicCS16}. Another overlooked aspect is that of \textit{intent}. As shown in Figure \ref{fig:abs}, we postulate that the noise can be further classified as intended and unintended noises. In the example, the adversarial perturbations, which are intentionally added, and the unintentional noise, such as Gaussian and salt \& pepper noises, have a similar effect, that is, changing the attribute label from male to female. Existing research has shown that several unintentional noise additions or corruptions can affect the original decision countering the detection mechanisms in place \cite{DBLP:conf/aaai/GoswamiRASV18,DBLP:conf/iclr/HendrycksD19,DBLP:conf/eccv/ModasROMF22}. Therefore, it is our understanding that we should not only be able to detect the unintended noises, but the approach should be able to disentangle intended adversarial perturbations from unintended noise patterns as well.  

In this paper, we present a detector network called CIAI detector which uses a Vision Transformer~\cite{DBLP:conf/iclr/DosovitskiyB0WZ21} modified with detection layers. 
It is trained in two stages using a novel Maximum Mean Discrepancy (MMD)~\cite{DBLP:journals/jmlr/GrettonBRSS12} and center-based loss along with the standard cross-entropy loss. CIAI can detect noises without having to consider the object classes and irrespective of the architecture used to craft the noises. We evaluate the results on CelebA~\cite{DBLP:conf/iccv/LiuLWT15}, ClelebA-HQ~\cite{DBLP:conf/iclr/KarrasALL18}, LFW~\cite{Huang2007a}, and AgeDB~\cite{DBLP:conf/cvpr/MoschoglouPSDKZ17} datasets for gender attribute prediction, CIFAR-10~\cite{krizhevsky2009learning}, and CIFAR-10-C~\cite{DBLP:conf/iclr/HendrycksD19} for 10-class classification. Further, attention maps and tSNE plots are used to indicate what the detection network focuses on for differentiating between original and modified images. 

\section{Related Work}
\paragraph{Attacks and Corruptions:} Adding some crafted and imperceptible noise can mislead an image classifier. In that regard, several gradient-based attacks have been proposed in the literature. FGSM (Fast Gradient Sign Method) \cite{DBLP:conf/iclr/WongRK20} is one of the first and fastest attacks that showcased classifiers' vulnerability to adversarial attacks. There are white-box attacks wherein the attacker has entire information regarding the model under attack, from parameters to training data, while in the black-box setting, the attacker has no information regarding the target model with the gray-box setting lying somewhere in between. FGSM is a single step $L_{\infty}$-distance-based attack. Under similar family falls attacks like PGD \cite{DBLP:conf/iclr/MadryMSTV18} which is iterative in nature, BIM \cite{DBLP:conf/iclr/KurakinGB17a}, RFGSM \cite{DBLP:conf/iclr/TramerKPGBM18}, MIFGSM \cite{DBLP:conf/cvpr/DongLPS0HL18}, SINIFGSM \cite{DBLP:conf/iclr/LinS00H20}, and so on. Based on $L_2$-distance, some other attacks proposed are CW \cite{DBLP:conf/sp/Carlini017}, DeepFool \cite{DBLP:conf/cvpr/Moosavi-Dezfooli16}, Auto-Attack \cite{DBLP:conf/icml/Croce020a}, and so on. Under $L_0$-distance, some proposed attacks are OnePixel \cite{DBLP:journals/tec/SuVS19}, Pixle \cite{DBLP:conf/ijcnn/PomponiSU22}, SparseFool \cite{DBLP:conf/cvpr/ModasMF19}, and such. Other than these, there are physical, semantic-based, and patch attacks among others. 

This varied range of adversarial attacks substantiates the vulnerability of neural networks and also shows how simple boundary manipulation can lead to almost complete failure of the well-trained models. Moreover, these attacks are often imperceptible by humans and can also be easily transferred under the black-box setting \cite{DBLP:journals/access/MahmoodMRD22}. There are also universal attacks that require minimal or no knowledge of the model \cite{DBLP:conf/ijcai/ZhangBLKWK21}. The careful formulation of such attacks ensures a grave drop in the model's performance, however, there is another way to reduce the model's performance. It comes in the form of corruption. While, many times, the corruptions happen unknowingly, like compressing images for effective storage, sometimes they can be added deliberately like improving the brightness of the image. In this regard, a benchmark \cite{DBLP:conf/iclr/HendrycksD19} was provided for the ImageNet dataset as ImageNet-C and ImageNet-P including 15 different corruptions at five different severity levels.       

\paragraph{Defense and Detection:} Be it adversarial attacks or corruptions, it is ideal to detect and defend against these modifications \cite{DBLP:journals/todaes/DongZ24,DBLP:journals/ijcv/GoswamiARSV19,DBLP:conf/iclr/MetzenGFB17,DBLP:journals/corr/abs-2312-09481,DBLP:conf/ndss/Xu0Q18}. Several defense and detection techniques have been proposed like adversarial training, distillation, and feature squeezing among others. We focus on detection techniques where the goal is to detect the modified images before sending them to the model. It can be done in a white-box setting, that is training the detector on a known attack and evaluating the detector for the specific attack, or in a black-box setting, where a trained detector should be able to detect images for an attack it has not seen before. Nearest neighbors and graph methods have been explored for such techniques \cite{DBLP:conf/iccv/AbusnainaWAWWYM21,DBLP:conf/cvpr/CohenSG20,DBLP:conf/nips/HuYGCW19}. 

Distribution-based methods have also been proposed like Local Intrinsic Dimensionality (LID) \cite{DBLP:conf/iclr/Ma0WEWSSHB18} that uses a Logistic Regression-based detector and Mahalanobis Distance (MD) \cite{DBLP:conf/nips/LeeLLS18}. MultiLID \cite{DBLP:conf/visapp/LorenzKK23}, built on LID, was recently proposed as a white-box detector that shows almost perfect detection when evaluated for binary classification between original and attacked images using non-linear classifiers 
Some techniques \cite{DBLP:conf/iclr/LiangLS18,DBLP:conf/nips/SunGL21} are specifically crafted for out-of-distribution detection. While these techniques can be extended for face datasets, more recently, an adversarial face detection \cite{DBLP:conf/ijcai/WangXLZLLCY23} was proposed using self-perturbations with a focus on GAN-based attacks as well. Almost all these techniques perform a binary classification to differentiate between unattacked and attacked images. Our aim is to be able to add another dimension, either in the form of unintentional noises (corruptions) or another family of attacks, to perform more than a binary classification.

\begin{figure*}[t]
\begin{center}
   \includegraphics[width=0.8\linewidth]{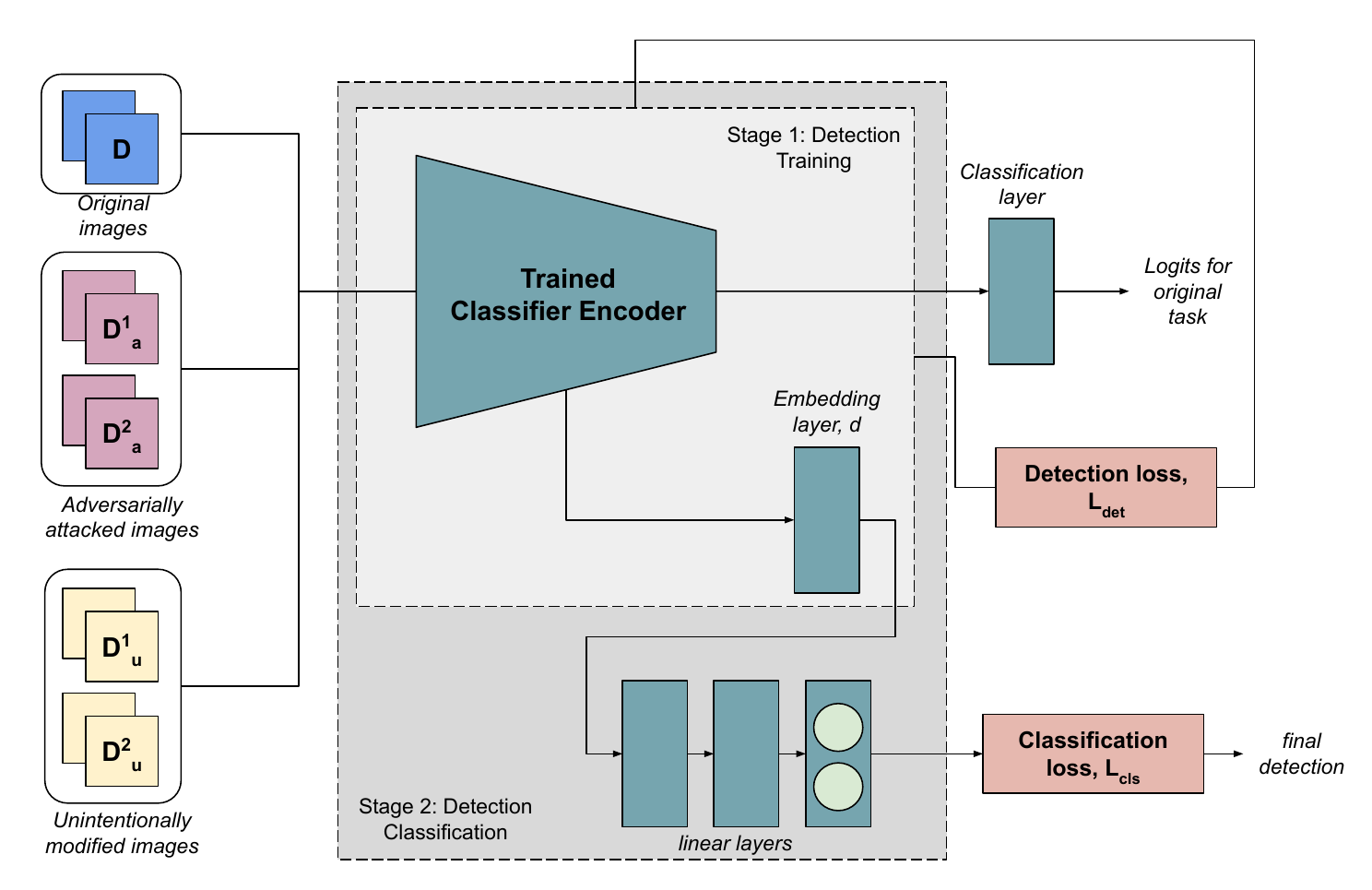}
\end{center}
   \caption{The proposed CIAI detection network (built on the trained classifier). The five image sets are taken from original images, images attacked using two different adversarial attacks, and images modified using two different unintentional noises.}
\label{fig:network}
\end{figure*}

\section{Proposed CIAI Network}
The CIAI network is proposed in this section considering adversarial perturbations as well as unintentional noises that can change the predicted attributed from one class to another like one gender label to another in the case of gender prediction task as seen in Figure \ref{fig:abs}. The inspiration is to be able to divide the modified images based on their distribution space in a way that a similar group of noises, whether seen or unseen, can be detected. We also use vision transformers for the detector with the understanding that they are known to generalize better than the CNNs \cite{DBLP:conf/cvpr/ZhangZZJZCZL022}.   

\textbf{With the Intent to Attack:} Detecting unintentional noises can help understand different adversarial attacks and the way they are designed. These unintentional noises do not affect the original accuracy of classifiers as much as the adversarial noises, but they still have an effect, and correcting the classifiers for such unintentional noises while discarding the intentionally attacked images can make the models more robust. These noises can occur at various stages while procuring the images or processing them including blurring and compression. Several of these unintentional noises are also used as data augmentation techniques, however, the idea here is to be able to detect the noises that lead to misclassification.  
The intended noises require a classifier to attack the images as they are crafted with the aim to fool the classifier but the unintended noises need no such network. They can be added by modifying the underlying distribution only slightly, like adding Gaussian noise, and still affect the model's performance \cite{DBLP:conf/cvpr/0001V0R20}. We indicate the concept of intent on gender attribution and classification tasks using these unintentional noises. The same can be further used for multi-class classification as well with a group of different occurring noises or using two different families of unintended corruptions. 

\subsection{Proposed CIAI Detection Network}
As shown in Figure \ref{fig:network}, we propose a Class-Independent Adversarial Intent (CIAI) detection network. The training process is in two stages. For the first stage, detection training, we use a novel MMD and Center-based loss to train a Vision Transformer initialized with classifier weights trained for the original recognition task. MMD has been disputed to not be aware of adversarial noise, especially the CW attack as reported in \cite{DBLP:conf/sp/Carlini017}. However, \cite{DBLP:conf/icml/GaoLZ0L0S21} uses a deep-kernel-based MMD to show that it is aware of adversarial attacks when a specific kernel is used. We build based on this observation by extracting image embeddings from the trained classifier and building a detection model minimizing the proposed novel loss, $L_{det}$ between randomly initialized centers and image embeddings. For the second stage, detection classification, we use the trained detector for detection classification by freezing all the layers and adding three trainable layers for training. 

We have a training set containing original images, $D$; a set of training images modified using adversarial attacks, $D^i_{a}$ where $i$ is the number of adversarial attacks used during training; and a set of training images modified using unintentional noises, $D^i_{u}$ where $i$ is the number of type of unintentional noises used. For our experiments, we use $i=2$, that is, two types of adversarial attacks and two types of unintentional noises, for training the detection network, $M$. We have a paired example at the end for training where $< I_j, I^{1}_{aj}, I^{2}_{aj}, I^{1}_{uj}, I^{2}_{uj}>$ is the $j^{th}$ training example and $I_j \in D$, $ I^{1}_{aj} \in D^1_{a}$, $ I^{2}_{aj} \in D^2_{a}$, $ I^{1}_{uj} \in D^1_{u}$, and $ I^{2}_{uj} \in D^2_{u}$. Five centers are initialized, one for each set of training images, $C_k$ where $k=5$ for our experiments with $k=0$ for $D$, $k=1$ for $D^1_{a}$, $k=2$ for $D^2_{a}$, $k=3$ for $D^1_{u}$, and $k=4$ for $D^2_{u}$. Further, for computing the loss, the detector model, $M$ is used to get the embeddings for each batch of images leading to $<E_j, E^{1}_{aj}, E^{2}_{aj}, E^{1}_{uj}, E^{2}_{uj}>$ where $M(I_j)$ gives the embedding for $I_j$ as $E_j$ and so on for all the other batches. 

\subsection{Original vs Modified Images}
The first loss is formulated based on dividing the original image space and modified image space, that is, dividing $D$ and $D^i_{a}$ as well as $D$ and $D^i_{u}$. We use MMD values between the initialized center and a batch of training images. For original images and modified images, we formulate the loss term, $L_1$ with three subterms, $L^{close}_{1}$, $L^{farorg}_{1}$ and $L^{farmod}_{1}$. 
\begin{equation}
\begin{aligned}
L^{close}_{1} &= MMD(C_0, E_j) + MMD(C_1, E^{1}_{aj}) \\ 
& + MMD(C_2, E^{2}_{aj}) + MMD(C_3, E^{1}_{uj}) \\ 
& + MMD(C_4, E^{2}_{uj})
\label{eq:l1close}
\end{aligned}
\end{equation}
The formulation for $L^{close}_{1}$ is based on bringing each center close to its image counterpart, as depicted in Equation \ref{eq:l1close}. The MMD uses each center as one distribution and each batch of images as the other, calculated with the aim of minimizing the distance between the two.   
 \begin{equation}
\begin{aligned}
L^{farorg}_{1} &= MMD(C_1, E_j) + MMD(C_2, E_j) \\
&+ MMD(C_3, E_j) + MMD(C_4, E_j)
\label{eq:l1org}
\end{aligned}
\end{equation}

The formulation for $L^{farorg}_{1}$ is based on bringing each center other than the one dedicated for $D$, that is, $C_k$ where $k=1,2,3,4$ away from the original images $I_j$, as depicted in Equation \ref{eq:l1org}. The MMD is calculated between each center and original images with the aim of maximizing the distance between the two.  
\begin{equation}
\begin{aligned}
L^{farmod}_{1} &= MMD(C_0, E^{1}_{aj}) + MMD(C_0, E^{2}_{aj}) \\
&+ MMD(C_0, E^{1}_{uj}) + MMD(C_0, E^{2}_{uj})
\label{eq:l1far}
\end{aligned}
\end{equation}

The formulation for $L^{farmod}_{1}$ is based on bringing the center dedicated for $D$, that is, $C_0$ away from the other set of images, that is, $I^{1}_{aj}$, $I^{2}_{aj}$, $I^{1}_{uj}$, and $I^{2}_{uj}$, as depicted in Equation \ref{eq:l1far}. The MMD is calculated between the center and each set of images other than the original images with the aim of maximizing the distance between the two. 
\begin{equation}
\begin{aligned}
L_1 = \alpha \times L^{close}_{1} - (1-\alpha) \times (L^{farorg}_{1} + L^{farmod}_{1})
\label{eq:l1}
\end{aligned}
\end{equation}
Equation \ref{eq:l1} depicts the entire loss term for creating a division between the original images and the other modified images. 

\subsection{Intentionally Modified vs Unintentionally Modified Images}
The next loss term $L_2$ is based on creating a separation between the intentionally modified and unintentionally modified images. The embeddings for different image settings are pulled apart for the detector network M and are formulated using two subterms, $L^{close}_{2}$ and $L^{far}_{2}$. 
\begin{equation}
\begin{aligned}
L^{close}_{2} &= MMD(C_1, E^{1}_{aj}) + MMD(C_2, E^{2}_{aj})\\ 
& + MMD(C_3, E^{1}_{uj}) + MMD(C_4, E^{2}_{uj})  
\label{eq:l2close}
\end{aligned}
\end{equation}

$L^{close}_{2}$, as presented in Equation \ref{eq:l2close} is meant for closing the distance between the image batches with their respective centers. That is, the embedding batch from the first adversarially attacked images, $E^{1}_{aj}$, is pulled close to its respective assigned center, $C_1$ by calculating the MMD value between the two. This is done for each set of embeddings with their respective center. Since this distance needs to be minimized, the term is added as a positive factor in the entire loss formulation. 
\begin{equation}
\begin{aligned}
L^{far}_{2} &= MMD(C_1, E^{1}_{uj}) + MMD(C_2, E^{2}_{uj})\\ 
& + MMD(C_3, E^{1}_{aj}) + MMD(C_4, E^{2}_{aj})
\label{eq:l2far}
\end{aligned}
\end{equation}
On the other hand, $L^{far}_{2}$ is used for creating distance between the adversarially attacked images and unintentionally modified images. For that, the embeddings from the first unintentionally modified images are pulled towards the center assigned for the first adversarially attacked images and vice-versa. This is further done for the second attack and unintentional noise as well, as depicted in Equation \ref{eq:l2far}. Since the centers are supposed to be pulled far from the embeddings in this formulation, the distance needs to be maximized here. 
\begin{equation}
\begin{aligned}
L_2 = \alpha \times L^{close}_{2} - (1-\alpha) \times L^{far}_{2} 
\label{eq:l2}
\end{aligned}
\end{equation}

The two terms $L^{close}_{2}$ and $L^{far}_{2}$ are then combined as shown in Equation \ref{eq:l2} with $\alpha$ as the regularization factor. This term can not only be used to divide the intended and unintended noises apart but also can be used between two families of intended attacks or two families of unintended attacks as shown in experiments in the coming sections. 

\subsection{Groups within Modified Images}
The third loss term, $L_3$ is used to create a separation within the subgroups of intentionally and unintentionally modified images. Since we use two types of noises in each group, a separation is created between these two noises. If we use FGSM and PGD attacks for the intentionally modified group, $L_3$ loss is used to pull the two groups away from each other. They can be further modified or removed depending on the number of noises used in each group. The loss is formulated using two subterms, $L^{close}_{3}$ and $L^{far}_{3}$. $L^{close}_{3}$, is formulated with the aim to bring the embedding of different image batches closer to their respective centers. Mathematically, the equation is the same as $L^{close}_{2}$ as seen in Equation \ref{eq:l2close}. 


\begin{equation}
\begin{aligned}
L^{far}_{3} &= MMD(C_1,E^{2}_{aj}) + MMD(C_2, E^{1}_{aj})\\ 
& + MMD(C_3, E^{2}_{uj}) + MMD(C_4, E^{1}_{uj})  
\label{eq:l3far}
\end{aligned}
\end{equation}

$L^{far}_{3}$, as shown in Equation \ref{eq:l3far}, is formulated to maximize the distance between embeddings from one attack from the center of another attack within the same group. That is, the images from the first adversarial attack are pulled away from the center dedicated to the second adversarial attack, and vice-versa.  
\begin{equation}
\begin{aligned}
L_3 = \alpha \times L^{close}_{3} - (1-\alpha) \times L^{far}_{3} 
\label{eq:l3}
\end{aligned}
\end{equation}
Finally, $L_3$ is combined using the two subterms as shown in Equation \ref{eq:l3} with $\alpha$ as the regularization term. All three loss terms together form the complete loss used to train the detector network, M. 

\subsection{Detector Network}

The detector network, M outputs an embedding and is trained using the detection loss $L_{det}$. The network is similar to the original classifier in that it is initialized with the weights of the trained classifier. The network is then modified by changing the last classifier layer to get the embeddings with dimension d seen as stage 1 of detection training.

\begin{equation}
L_{det} = \beta \times L_1 + \gamma \times L_2 + \delta \times L_3
\label{eq:detloss}
\end{equation}

The formulation for $L_{det}$ is depicted in Equation \ref{eq:detloss}, where $\beta + \gamma + \delta = 1$.  The CIAI network, once trained, is then modified by adding 3 linear layers to train for 2-class, 3-class, or 5-class detection as depicted in Figure \ref{fig:network} seen as stage 2 of detection training. For the 2-class detector, the detection is between original and modified images; for the 3-class detector, the detection is between the original, modified with attacks, and modified with unintentional noises. The 5-class detector is for detection between original images, two types of adversarial attacks, and two types of unintentional attacks.  

\section{Experiments}
For detection, the results are shown and discussed for different face datasets and the CIFAR dataset~\cite{krizhevsky2009learning} (see supplementary), where classifiers are trained for attribute prediction and classification tasks, respectively. For the experiments, we first train the classifier for original tasks and then use them to train the proposed CIAI detection network. We use additional datasets to see how the CIAI detector performs across different face datasets. We then present the results for detection along with the tSNE plots with an attention map analysis.  

\begin{figure}[]
\begin{center}
   \includegraphics[width=0.95\linewidth]{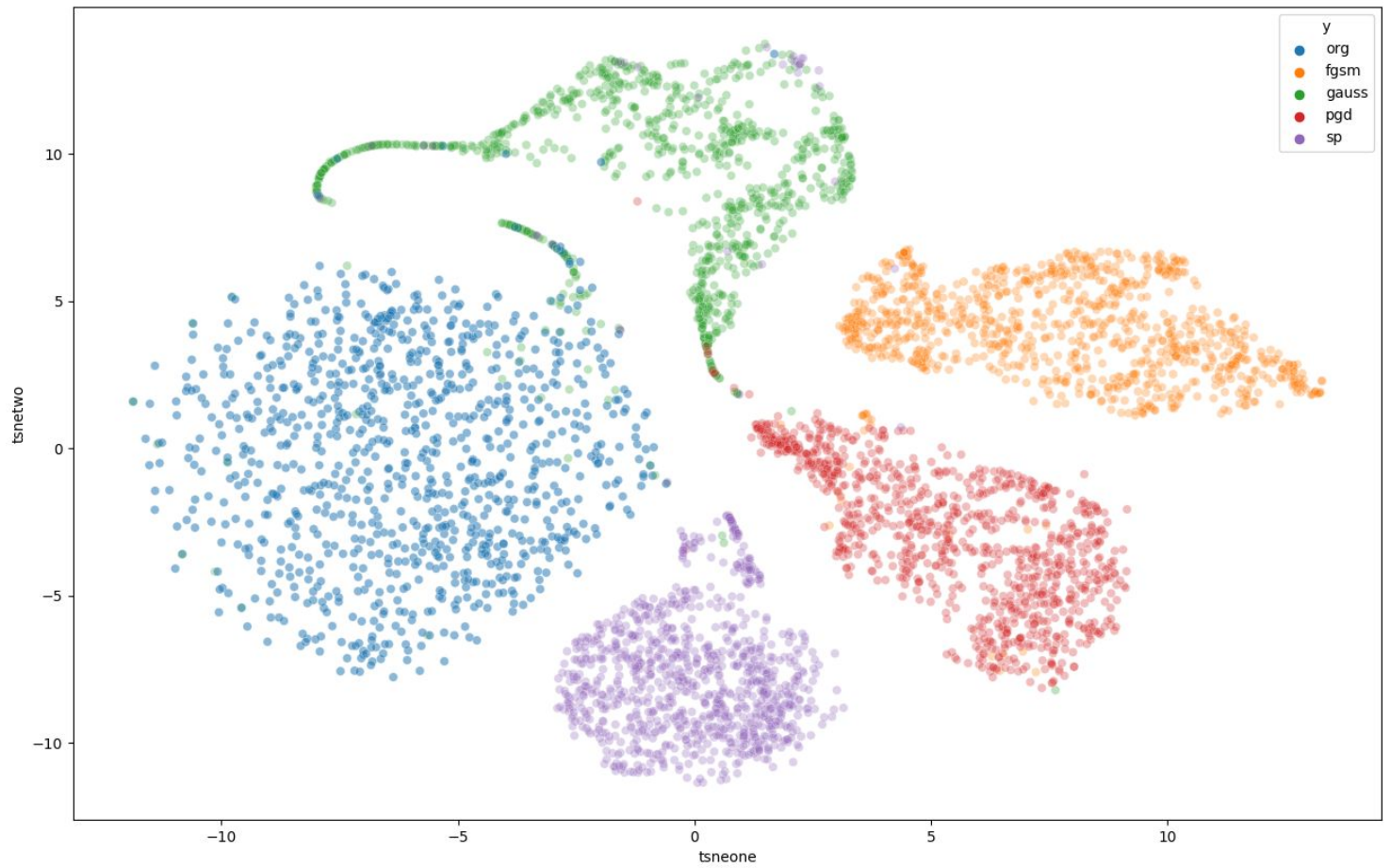}
\end{center}
   \caption{tSNE Plot for the proposed CIAI Detector trained on CelebA~\cite{DBLP:conf/iccv/LiuLWT15} dataset for gender prediction.}
\label{fig:celeb}
\end{figure}

\begin{figure}[]
\begin{center}
   \includegraphics[width=0.95\linewidth]{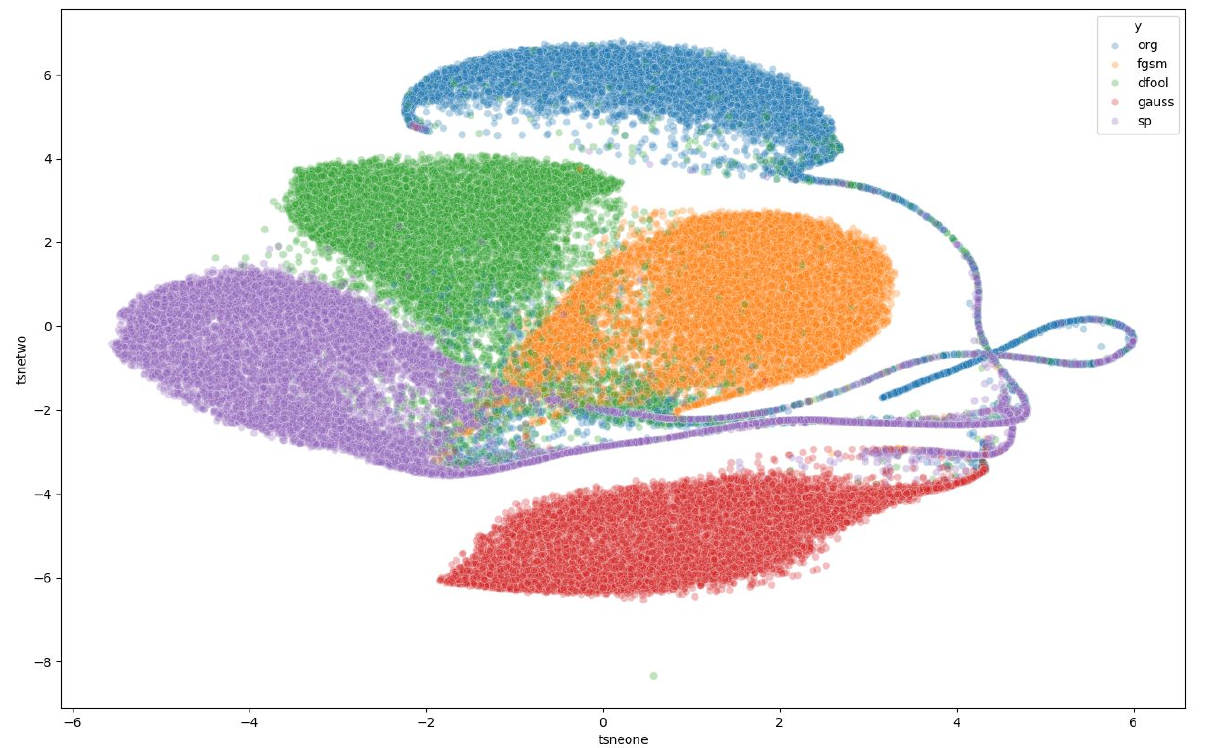}
\end{center}
   \caption{tSNE Plot for the proposed CIAI Detector trained on LFW~\cite{Huang2007a} dataset for gender prediction.}
\label{fig:lfw}
\end{figure}

\begin{table*}[]
\begin{center}
\begin{tabular}{|l|c|cccc|cccc|}
\hline
    & & \multicolumn{4}{c|}{Seen Attacks and Noises} & \multicolumn{4}{c|}{Unseen Attacks and Noises} \\ \cline{2-10}
       & Org & FGSM & PGD & Gaussian & SP & FFGSM & RFGSM & BIM & Speckle\\ \hline
Class acc & 99.58                   & 4.90                      & 0.42                    & 97.80                         & 98.12                  & 5.67                      & 0.38                      & 0.02      & 97.80              \\
$CIAI_{cel}$ (ours)   & 99.52                   & 99.98                    & 99.5                    & 91.33                        & 99.63                  & 99.95                     & 99.93                     & 99.92    & 94.30\\ \hline              
\end{tabular}
\caption{Classification accuracy (gender attribute prediction) and manipulated image detection results on the CelebA dataset for a 3-class classification setting. The CIAI detector is trained using FGSM and PGD as adversarial attacks and Gaussian and Salt \& Pepper noise as unintentional noises. The remaining (FFGSM, RFGSM, BIM, and Speckle) are results evaluated on unseen attacks and noises.} 
\label{tab:celeb}
\end{center}
\end{table*}

\begin{table*}[]
\begin{center}
\begin{tabular}{|l|c|cc|cc|cc|}
\hline
& & \multicolumn{2}{c|}{Seen Attacks} & \multicolumn{2}{c|}{Seen Noises} & \multicolumn{2}{c|}{Unseen Attacks} \\ \cline{2-8}
            & Org   & FGSM  & DeepFool & Gaussian & SP    & PGD   & CW    \\
            \hline
Cl acc      & 97.66 & 3.92  & 1.37     & 97.45    & 97.34 & 0.10  & 2.96  \\
$CIAI_{lfw}$ (ours) & 91.88 & 99.25 & 90.13    & 99.89    & 94.27 & 95.14 & 84.20 \\ \hline
\end{tabular}
\caption{Classification accuracy (gender attribute prediction) and manipulated image detection results on the LFW dataset for a 3-class classification setting. The CIAI detector is trained using FGSM and DeepFool as adversarial attacks and Gaussian and Salt \& Pepper noise as unintentional noises. The remaining (PGD and CW) are results evaluated on unseen attacks.} 
\label{tab:lfw}
\end{center}
\end{table*}

\subsection{Experimental Setting and Implementation Details}

To showcase the workings of the proposed approach, we have used two case studies: (a) gender prediction using face images and (b) a standard image recognition task. CelebA~\cite{DBLP:conf/iccv/LiuLWT15}, CelebA-HQ~\cite{DBLP:conf/iclr/KarrasALL18}, LFW~\cite{Huang2007a}, AgeDB~\cite{DBLP:conf/cvpr/MoschoglouPSDKZ17}, and CIFAR-10~\cite{krizhevsky2009learning} datasets are used for the case studies here. 

\paragraph{Gender Prediction using CelebA and LFW Dataset:} The CelebA dataset contains 162,770 training images, 20,367 validation images, and 19,962 testing images. The gender prediction is done between the two reported genders, male and female. Each image is resized to $3\times224\times224$ for the transformer input. The classifier for attribute prediction is trained for 5 epochs on the entire training set with a learning rate of $1e-4$, giving a final classification value of $99.58\%$. The CIAI network is trained with an embedding dimension, $d = 128$. It is initialized with the weights from the trained classifier. $CIAI_{cel}$ is first trained for 3 epochs with a learning rate of $1e-4$. Further, the linear layers are added to the detection network, and with frozen layers for the entire network except for the linear layers, the network is trained for 2-class and 3-class classification for another 3 epochs at a learning rate of $1e-4$. Further, we also train the CIAI detector for the LFW dataset which contains 13,233 images of 5749 people. For the gender labels, we refer to labels provided by Afifi and Abdelhamed~\cite{DBLP:journals/jvcir/AfifiA19} and train the classifier on the two reported genders, male and female. With 4272 males and 1477 females, the dataset can be seen as gender imbalanced. We randomly split the data into 10,000 training images, 1144 validation images, and 1144 testing images. Just like the $CIAI_{cel}$ detector, the $CIAI_{lfw}$ detector is trained with the same training parameters at every step. Both 2-class and 3-class detectors are trained for the LFW dataset. 

\begin{table}[]
\begin{center}
\begin{tabular}{|l|c|cc|cc|}
\hline
\multirow{2}{*}{Datasets $\downarrow$} & & \multicolumn{2}{c|}{Seen Attacks} & \multicolumn{2}{c|}{Seen Noises} \\ \cline{2-6}
      & Org   & FGSM  & PGD   & Gaussian & SP    \\ \hline
AgeDB & 99.22 & 100   & 99.90 & 79.95    & 98.44 \\
LFW   & 100   & 99.20 & 98.95 & 89.90    & 100  \\ \hline

\hline
\end{tabular}
\caption{Cross Dataset validation using the $CIAI_{cel}$ detector on AgeDB and LFW dataset in a 2-class setting.} 
\label{tab:cross-data}
\end{center}
\end{table}

\begin{table}[]
\begin{center}
\begin{tabular}{|l|cc|cc|}
\hline
\multirow{2}{*}{Algorithms $\downarrow$} & \multicolumn{2}{c|}{Seen Attacks} & \multicolumn{2}{c|}{Unseen Attacks} \\ \cline{2-5}
            & FGSM  & PGD   & BIM   & RFGSM \\ \hline
LID         & 76.70 & 70.70 & 74.0  & 73.00 \\
SID         & 99.70 & 73.70 & 81.80 & 77.80 \\
ODIN        & 75.60 & 71.60 & 71.10 & 75.20 \\
ReAct       & 92.30 & 88.40 & 89.20 & 89.10 \\
SPert       & \textbf{100}   & \textbf{100}   & \textbf{100} & \textbf{100}   \\ 
$CIAI_{lfw}$ (Ours) & \textbf{100} & \textbf{100} & \textbf{100}   & 99.70 \\
\hline
\end{tabular}
\caption{Detection accuracy across different methods including our CIAI in a 2-class setting trained on FGSM and PGD as seen attacks on the LFW dataset.} 
\label{tab:lfw-compare}
\end{center}

\begin{center}
\begin{tabular}{|l|cc|cc|}
\hline
\multirow{2}{*}{Algorithms $\downarrow$} & \multicolumn{2}{c|}{Seen Attacks} & \multicolumn{2}{c|}{Unseen Attacks} \\ \cline{2-5}
            & FGSM  & PGD   & BIM   & RFGSM \\ \hline
LID         & 82.00 & 52.50 & 55.20 & 54.40 \\
SID         & 96.70 & 63.40 & 79.20 & 72.80 \\
ODIN        & 76.70 & 75.80 & 75.70 & 75.70 \\
ReAct       & 93.60 & 90.50 & 90.70 & 90.60 \\
SPert       & \textbf{99.70} & \textbf{99.60} & 99.00 & 99.40 \\ 
$CIAI_{cel}$ (Ours) & 99.60 & 99.40 & \textbf{99.50} & \textbf{99.50} \\
\hline
\end{tabular}
\caption{Detection accuracy of the existing and proposed CIAI (trained on CelebA dataset) in a 2-class setting trained on FGSM and PGD as seen attacks. The results are on the CelebA-HQ dataset.} 
\label{tab:chq-compare}
\end{center}
\end{table}

\begin{figure*}[]
\begin{center}
   \includegraphics[width=0.75\linewidth]{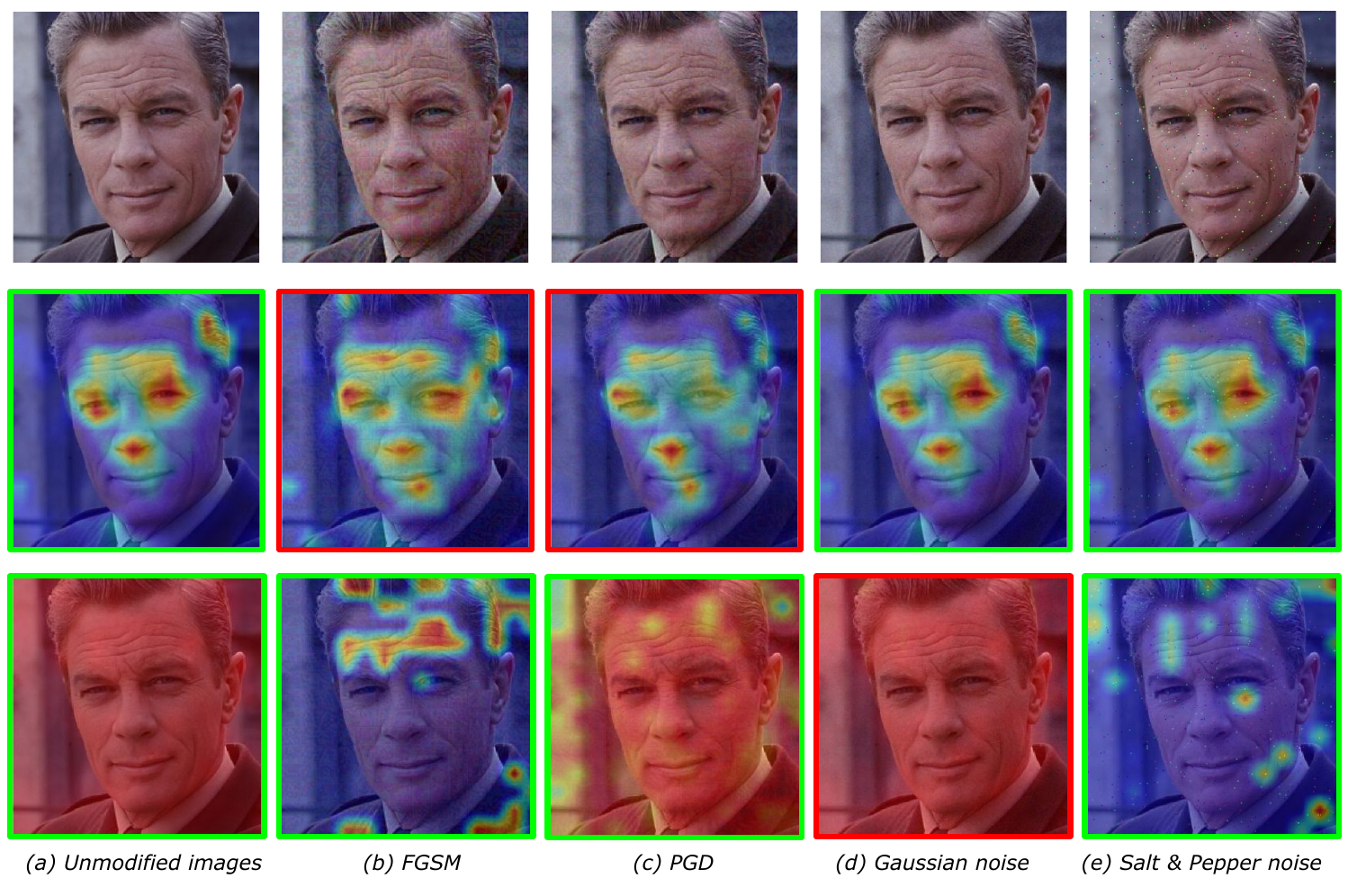}
\end{center}
   \caption{Attention Maps for the AgeDB dataset. The top row indicates the modified images used in the experiments. The middle row shows the attention maps for the attribute prediction task; the green box indicates the correct classification label, that is, male here, and the red box indicates the incorrect classification label, that is, female. The last row shows attention maps for the detection task to detect intentional and unintentional noises; the green box indicates correct classification in the 3-class setting while red indicates incorrect classification.}
\label{fig:attn-age}
\end{figure*}

\begin{figure*}[!t]
\begin{center}
   \includegraphics[width=0.75\linewidth]{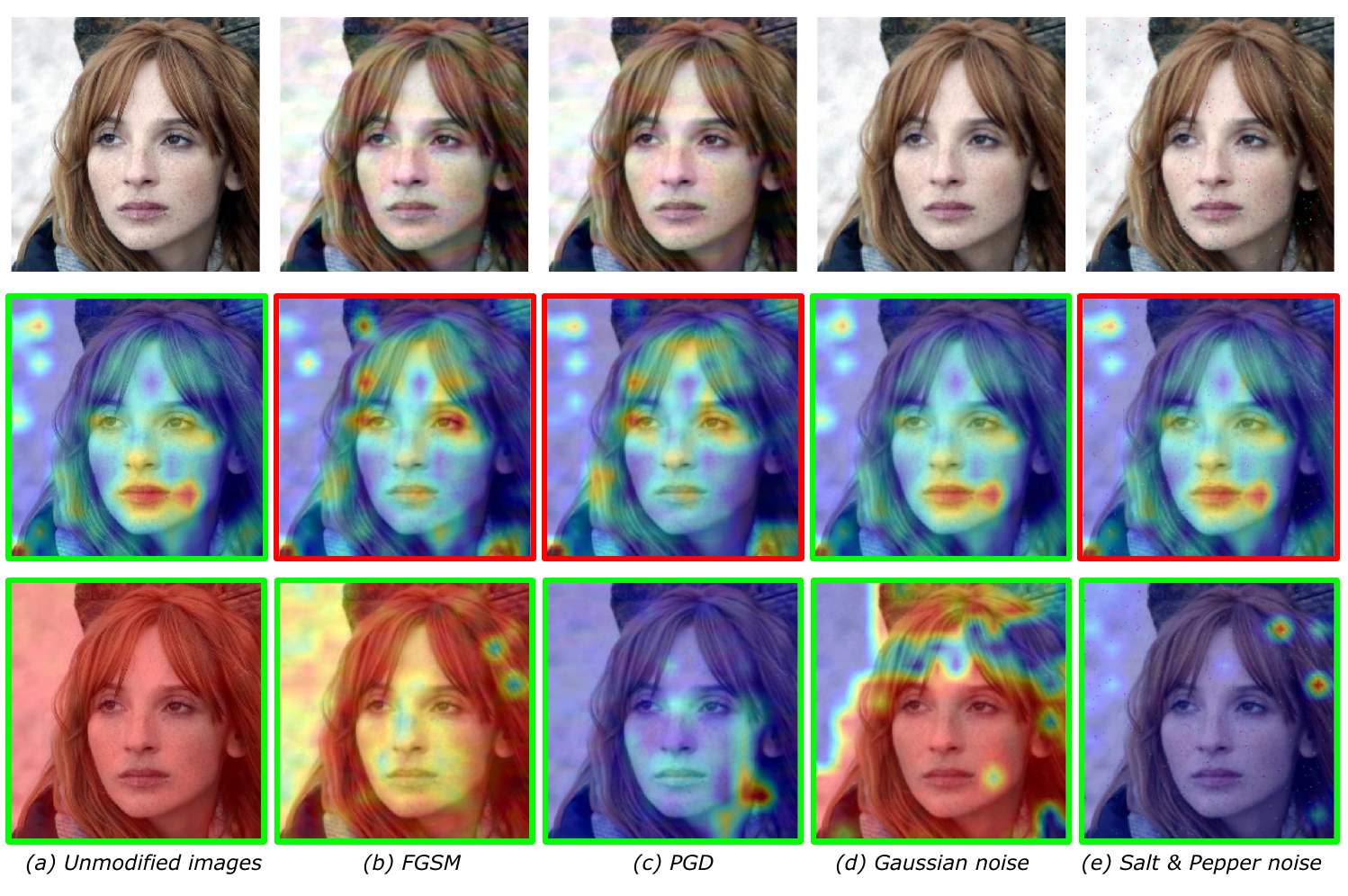}
\end{center}
   \caption{Attention Maps for the CelebA-HQ dataset. The top row indicates the modified images used in the experiments. The middle row shows the attention maps for the attribute prediction task; the green box indicates the correct classification label, that is, female here, and the red box indicates the incorrect classification label, that is, male. The last row indicates attention maps for the detection task to detect intentional and unintentional noises; the green box indicates correct classification in the 3-class setting while red indicates incorrect classification.}
\label{fig:attn-celebhq}
\end{figure*}

\paragraph{Cross Dataset Validation and other comparative results:} For evaluating the performance of the trained detectors across datasets, we use the AgeDB and LFW datasets. The dataset contains 12,240 images of 440 subjects with attribute information for identity, gender, and age. The entire dataset is used for validation on the CIAI detectors trained on the LFW dataset and CelebA dataset, respectively. Further, we use the CelebA-HQ dataset to compare against existing state-of-the-art detection methods. The CelebA-HQ dataset is a subset of the CelebA dataset with 30,000 high-quality images in totality, out of which 1000 images are a part of the testing set. For comparison, we use the CIAI detector trained on the CelebA dataset; the images included in the testing set of the CelebA-HQ dataset are intentionally removed from the training and validation set during the training of the attribute classifier as well as the CIAI detector. For comparison, we consider 5 best-performing methods: \textbf{LID \cite{DBLP:conf/iclr/Ma0WEWSSHB18}} or Local Intrinsic Dimensionality uses a k-nearest neighbor classifier for detecting adversarial attacks; \textbf{SID \cite{DBLP:conf/aaai/Tian0LD21}} utilizes wavelet transformation to detect the adversarial attacks; \textbf{ODIN \cite{DBLP:conf/iclr/LiangLS18}} is specifically designed for detecting out-of-distribution examples; \textbf{ReAct \cite{DBLP:conf/nips/SunGL21}} is a post hoc method, also proposed for out-of-distribution detection and rectifies the internal activations of the neural networks; \textbf{SPert \cite{DBLP:conf/ijcai/WangXLZLLCY23}} uses original datasets with their self-perturbations to train a detector.

\paragraph{Seen and Unseen Noises:} For the experiments, a number of attacks and noises are used to modify images. Seen noises are noises used to train the detector, and unseen noises are used for evaluation. For the \textbf{CelebA dataset}, FGSM \cite{DBLP:journals/corr/GoodfellowSS14}, and PGD \cite{DBLP:conf/iclr/MadryMSTV18} are used as seen adversarial attacks, and Fast FGSM \cite{DBLP:conf/iclr/WongRK20}, RFGSM \cite{DBLP:conf/iclr/TramerKPGBM18}, and BIM \cite{DBLP:conf/iclr/KurakinGB17a} as unseen adversarial attacks. On the other hand, Gaussian and salt \& pepper noise (also known as impulse noise) are used as seen unintended noises, and speckle as unseen unintended noises. All these noises affect the gender prediction accuracy (Table \ref{tab:celeb}). For the \textbf{LFW dataset}, the seen adversarial attacks are FGSM and DeepFool \cite{DBLP:conf/cvpr/Moosavi-Dezfooli16}, while the unseen attacks are PGD and CW. For seen unintended noises, Gaussian and salt \& pepper noises are used. 

\subsection{Detection Results on CelebA Dataset}
For, $L_{det}$ (Equation \ref{eq:detloss}), the values used for $\beta$, $\gamma$, and $\delta$ is 
$1/3$, $1/3$, and $1/3$, and the regularization value $\alpha = 0.3$. The trained classifier is used to create adversarially attacked images for FGSM and PGD attacks. For unintentional noises, only the images are required, with no classifier. Gaussian and Salt \& Pepper noises are used for unintentional noises. After stage 1, the tSNE plot for 500 randomly selected test images is plotted as seen in Figure \ref{fig:celeb}. As seen in the plot, the CIAI network divided the five groups quite distinctively, which can be further substantiated by the high detection accuracy. All the results are shown on the testing set of 19,962 images. The detection results are further depicted in Table \ref{tab:celeb}, where the original images and adversarially attacked images can be detected with almost perfect accuracy. Gaussian and salt \& pepper noise can be also detected with great accuracy even when the classification accuracy remains quite high.  

\subsection{Detection Results on LFW Dataset}
With the same training parameters as the detector trained on the CelebA dataset, the CIAI detector for the LFW dataset is trained, and all the results are shown on the testing set of 1144 images (Table \ref{tab:lfw}). After the stage 1 pretraining, the tSNE plots for the five variations - original images and images modified by FGSM, DeepFool, Gaussian, and Salt \& Pepper noises as seen in Figure \ref{fig:lfw}. As seen in the plot, the network divides the majority of the images quite distinctively. Empirically as well, the detection is done with a high accuracy.  

\subsection{Cross Dataset Validation and Comparisons}
For sets of experiments reported in Tables 3 to 5, we use a 2-class setting for the CIAI detector to differentiate between original and adversarial images. For the cross-dataset validation, we use the CIAI detector trained on the CelebA dataset, $CIAI_{cel}$, and test it on the entire AgeDB dataset as well as the testing set for LFW dataset, as reported in Table \ref{tab:cross-data}. The detector gives high detection accuracy for all noises. We can conclude that the detector performs well, even when the data distribution changes. Further, we compare our detector with other existing methods as seen in Table \ref{tab:lfw-compare} for the LFW dataset and Table \ref{tab:chq-compare} for the CelebA-HQ dataset. The detector's performance is comparable to the best-performing detector.  

\subsection{Attention Map Analysis}
For attention map analysis, attention weights are pulled from the last multi-head attention layer of the Transformer classifier and detector. The first row in Figures \ref{fig:attn-age} and \ref{fig:attn-celebhq} indicate the original unmodified image along with the modified images. Further, attention maps for the attribute predictor, as shown in the middle row, indicate the features utilized to classify the images and how different noises affect the decision. Even with the unintended noises, the attention changes slightly, and empirically, the confidence level decreases during attribute prediction even if there is no misclassification. For both examples, we can see that the FGSM and PGD adversarial attacks successfully fool the classifier by predicting the wrong gender label. For Gaussian noise, in Figure \ref{fig:attn-age}, the modification fails to change the label and the detector also fails to identify the noise. However, we see a 25\% decrease in the confidence level after the noise is added for the classification. We can also observe that the detector considers the entire image when detecting the unmodified images but focuses on particular regions when detecting intentional and unintentional noises as seen in the last row of two figures. The detector is thus able to learn the difference between these modifications. 

\section{Conclusion}
In this paper, we introduced CIAI, a novel noise detection network that operates independently of the image class. CIAI not only distinguishes between original and modified images but also differentiates between intentional (adversarial) and unintentional noise, both of which can impact the performance of a model. Our results show that CIAI performs effectively on both known and unknown noise types, including those with similar characteristics. When $L_p$-norm-based attacks are used as seen noises, attacks based on similar formulations are detected with almost similar accuracy. As observed when FGSM is the seen noise, and FFGSM and RFGSM are unseen noises for the CelebA dataset. Additionally, it can be tailored to specifically target adversarial attacks. Importantly, our findings demonstrate that CIAI maintains robust detection capabilities even when classification accuracy is not significantly compromised by unintentional noise.

\section{Acknowledgement}
The research is supported by a grant from iHub-Drishti, Technology Innovation Hub (TIH) at IIT Jodhpur. 



{\small
\bibliographystyle{ieee}
\bibliography{main}
}

\end{document}